\documentclass[runningheads]{llncs}
\usepackage{graphicx}
\usepackage{inconsolata}
\usepackage{algorithm}
\usepackage{algorithmic}
\usepackage{amssymb}
\usepackage{dsfont}
\usepackage{amsmath}
\usepackage{booktabs}
\usepackage{multirow}
\usepackage[normalem]{ulem}
\useunder{\uline}{\ul}{}
\usepackage{amssymb}
\usepackage{bbm}
\usepackage{graphicx}
\usepackage{CJKutf8}
\usepackage[backref]{hyperref}
\begin{document}
\title{GeoGLUE: A GeoGraphic Language Understanding Evaluation Benchmark}
%

\author{Dongyang Li\inst{1} \and
Ruixue Ding\inst{2} \and
Qiang Zhang\inst{2} \and
Zheng Li\inst{2} \and
Boli Chen\inst{2} \and
Pengjun Xie\inst{2} \and
Yao Xu\inst{2} \and
Xin Li\inst{2} \and
Ning Guo\inst{2} \and
Fei Huang\inst{2} \and
Xiaofeng He\inst{1}}
\authorrunning{Dongyang Li, Ruixue Ding et al.}
%
\institute{East China Normal University, Shanghai, China\\
\email{dongyangli0612@gmail.com, hexf@cs.ecnu.edu.cn}\\ \and
Alibaba Group, Hangzhou, China\\
\email{\{ada.drx, muxi.zq, hengchong.lz, boli.cbl, chengchen.xpj, xuenuo.xy, beilai.bl, ning.guo, f.huang\}@alibaba-inc.com}}

\maketitle              
\begin{abstract}
With a fast developing pace of geographic applications, automatable and intelligent models are essential to be designed to handle the large volume of information. However, few researchers focus on geographic natural language processing, and there has never been a benchmark to build a unified standard. In this work, we propose a \textbf{GeoG}raphic \textbf{L}anguage \textbf{U}nderstanding \textbf{E}valuation benchmark, named GeoGLUE. We collect data from open-released geographic resources and introduce six natural language understanding tasks, including geographic textual similarity on recall, geographic textual similarity on rerank, geographic elements tagging, geographic composition analysis, geographic where what cut, and geographic entity alignment. We also provide evaluation experiments and analysis of general baselines, indicating the effectiveness and significance of the GeoGLUE benchmark\footnote{The GeoGLUE benchmark is available at \url{https://modelscope.cn/datasets/damo/GeoGLUE/summary}}. 

\keywords{Benchmark  \and Geographic \and Natural Language Understanding.}
\end{abstract}

\section{Introduction}

\begin{figure}[t]
\centering
\includegraphics[width=0.8\columnwidth]{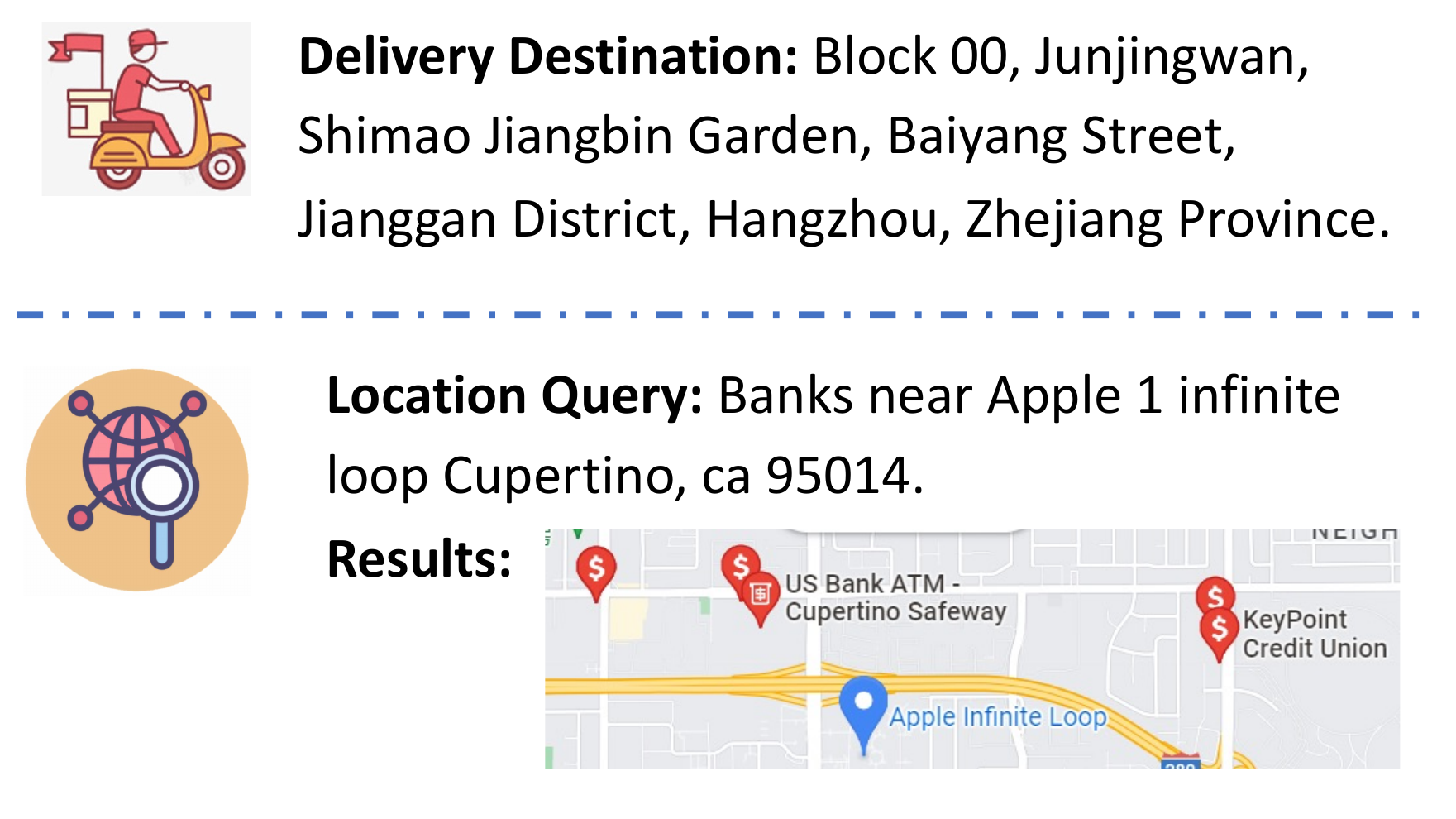}
\caption{Examples of geographic information usage scenarios. (Best viewed in color).}
\label{motivation}
\end{figure}

Geographic-based applications develop rapidly and appear throughout daily life. Logistics Distribution \cite{daugherty1998leveraging}, Surroundings Discovery \cite{DBLP:conf/gis/AydinGGCT13}, Position Navigation are the main daily usage scenarios. These applications generate a large volume of geographic data waiting for analysis and reutilization. Meanwhile, the characteristics of geographic data differ from other general domain expressions. Geographic data contains Geographic Information System \cite{antenucci1991geographic} (GIS)-based location knowledge for place positioning and the geographic text describes with geographic style expression (e.g., mixed usage of LA and Los Angeles, a special restaurant name called Drunk and Missed a Shoe Barbecue Restaurant, etc.) which requires particular processing. 
Fig. \ref{motivation} shows two typical geographic information usage scenarios, the top part of the figure indicates a delivery destination address involving a set of particular position words, and the bottom part of the figure presents a search query about banks and its relevant POIs\footnote{POIs are location points in maps or geo-datasets, which use icons as expressional forms and are defined mainly by geographical coordinates (i.e., longitude and latitude).} (Points Of Interest) results on the map.

The large volume of data generation leads Artificial intelligence (AI) models to manage the data with low costs. The problem of how to offer suitable evaluation criteria for different models needs to be addressed.
Previous works exploit benchmarks to unify the research criteria of various AI models. \cite{DBLP:conf/iclr/WangSMHLB19,DBLP:conf/nips/WangPNSMHLB19} propose general benchmarks for natural language understanding evaluations with several tasks and datasets. \cite{DBLP:journals/tacl/KwiatkowskiPRCP19,DBLP:conf/acl/NieWDBWK20} design benchmarks for specific tasks, question answering, and natural language inference. \cite{DBLP:conf/emnlp/RaganatoPCP20} and \cite{DBLP:conf/acl/RybakMTG20} are multi-lingual and low-resource language benchmarks. \cite{DBLP:conf/acl/ParcalabescuCMF22,DBLP:conf/emnlp/Pezzelle0GGB20} introduce multi-modal benchmarks for language-vision related methods. However, these works focus on the general domain and do not consider the specific-domain data, which has special domain features.

In specific domain, \cite{DBLP:conf/acl/ZhangCBLLSYTXHS22} concentrates on biomedical natural language understanding evaluation. \cite{DBLP:conf/acl/ChalkidisJHBAKA22} introduces a legal-based domain language processing benchmark. \cite{DBLP:conf/emnlp/BarbieriCAN20,DBLP:conf/emnlp/ElSheriefZMASCY21} design benchmarks for social media text. 
Little consideration is attached to the geographic domain.
Although \cite{DBLP:conf/naacl/LiLXL19,DBLP:conf/emnlp/HuangSLWCZDQ19} are geographically related works, they are only centered on single application scenarios and inadequate at generality.

To fill the gaps mentioned above in geographic benchmarks, we propose a \textbf{GeoG}raphic \textbf{L}anguage \textbf{U}nderstanding \textbf{E}valuation benchmark (GeoGLUE), which consists of six geographic text-related tasks, geographic textual similarity on recall, geographic textual similarity on rerank, geographic elements tagging, geographic composition analysis, geographic where what cut and geographic entity alignment. All tasks' datasets are collected from open-released resources, and we devote efforts to the data's privacy protections. Experiments of five strong baselines on our benchmark are conducted, and the detailed results' analyses demonstrate our work's effectiveness and profitableness.

Accordingly, our contributions can be summarized as follows:
\begin{itemize}
\item We propose the first-ever comprehensive geographic natural language understanding evaluation benchmark, named GeoGLUE, to settle the lack of geographic-domain specific evaluation works problem.
\item GeoGLUE comprises six commonly utilized natural language processing (NLP) tasks, and each corresponding dataset is collected from geographically related open resources.
\item We conduct multiple experiments on five baselines to report the effectiveness of our benchmark and its valuableness for further research.
\end{itemize}

\section{Related Work}
\subsection{General-Domain Benchmarks}
Recently, several works have been proposed to boost the development of the evaluation benchmark research process. We divide these works into three categories. 
(1) Task-based benchmarks.  
GLUE \cite{DBLP:conf/iclr/WangSMHLB19} is a general natural language benchmark, which is trained on plenty of datasets and contains nine existing tasks. 
SuperGLUE \cite{DBLP:conf/nips/WangPNSMHLB19} is proposed as an advanced version of GLUE via designing more challenging tasks and more diverse task formats. 
Natural Questions \cite{DBLP:journals/tacl/KwiatkowskiPRCP19} is designed for the particular question answering task and is compatible with long and short answers simultaneously.
\cite{DBLP:conf/acl/DeYoungJRLXSW20} introduces several metrics to measure the faithfulness of rationales provided by reasoning models.
\cite{DBLP:conf/acl/NieWDBWK20,DBLP:conf/conll/RavichanderNRH19} are proposed for natural language inference tasks and provide new challenge test sets.
(2) Language-based benchmarks. 
CLUE \cite{DBLP:conf/coling/XuHZLCLXSYYTDLS20} proposes a Chinese general language evaluation benchmark for single-sentence and sentence-pair understanding tasks. 
\cite{DBLP:conf/ijcnlp/WilieVWCLLSMFBP20,DBLP:conf/coling/KotoRLB20} introduce the specific Indonesian language benchmark with multiple different complexity tasks to fix the lack of available resources problem.
KLEJ \cite{DBLP:conf/acl/RybakMTG20} proposed a comprehensive multi-task benchmark for Polish language understanding, together with an online leaderboard. Likewise, \cite{DBLP:conf/emnlp/HamCPCS20} is for a Korean benchmark, \cite{DBLP:conf/emnlp/ShavrinaFESAMMT20} is for a Russian benchmark.
Meanwhile, \cite{DBLP:conf/emnlp/RaganatoPCP20,DBLP:journals/corr/abs-2003-11080} propose multi-lingual benchmarks, which are the capability of handling numbers of language representations. 
(3) Modal-based benchmarks.
\cite{DBLP:conf/acl/ParcalabescuCMF22,DBLP:conf/emnlp/Pezzelle0GGB20} introduces language and vision fine-grained evaluations benchmark for multi-modal task scenarios. 

However, these benchmarks' data are collected from the general-domain corpus and can be extensively used for general models.

\subsection{Specific-Domain Benchmarks}
In addition, there also exist multiple works focused on the specific domain. 
(1) Biomedical domain.
\cite{DBLP:conf/acl/ZhangCBLLSYTXHS22} establishes a Chinese Biomedical language understanding evaluation benchmark with eight tasks and a Leaderboard to promote the development of biomedical domain research.
\cite{DBLP:journals/corr/abs-2004-10706} proposes a biomedical dataset for further utilizations on natural language understanding tasks.
(2) Legal domain. 
\cite{DBLP:conf/acl/ChalkidisJHBAKA22} introduces various legal domain language processing tasks, leading to convenient employment for legal practitioners.
(3) Social media domain.
TweetEval \cite{DBLP:conf/emnlp/BarbieriCAN20} proposes seven heterogeneous Twitter-specific classification tasks to solve the bottleneck of corpus fragmentation and data deficiencies.
\cite{DBLP:conf/emnlp/ElSheriefZMASCY21} proposes a theoretical taxonomy of implicit hate speech and a benchmark with fine-grained data.
(4) Quotations domain. 
QuoteR \cite{DBLP:conf/acl/QiYYCLS22} introduces a three language based benchmark for quote recommendation methods to align the evaluation conditions.

Few works touch on the tasks of geographic natural language processing and are not comprehensive in terms of task coverage.
\cite{DBLP:conf/naacl/LiLXL19} pays attention to the address text parsing task. \cite{DBLP:conf/lrec/MandlGNFSSW08,DBLP:conf/emnlp/RollerSRWB12} concern geographic text documents, but pay less attention on GIS data. 
\cite{DBLP:conf/emnlp/LiKCC22} proposes a pre-trained language model which is trained on geographic data but does not release the dataset.
Thus, we introduce a more comprehensive geographic language understanding benchmark to facilitate domain development. 

\section{GeoGLUE Overview}
In order to promote the development of geographic natural language processing, we propose a geo-specific language understanding benchmark, which consists of six tasks covering searching, classification, and alignment fields.

\subsection{The Characteristics of GeoGLUE}
To ensure the quality of GeoGLUE, we adjust the benchmark to have the following characteristics.

\textbf{Reusability}
The benchmark publishing site provides users with high-quality documentation and detailed tutorials. 
We also develop a detailed and sustainable iterative update plan for this benchmark. 
GeoGLUE has been applied on a published paper \cite{DBLP:journals/corr/abs-2301-04283}. 
As of Due Date of the paper, GeoGLUE has been downloaded nearly 2,700 times, and the model trained with GeoGLUE's data has been downloaded nearly 294,000 times\footnote{\url{https://modelscope.cn/models?name=mgeo&page=1}}.

\textbf{Diversity}
The tasks of the GeoGLUE benchmark cover six geographic NLP scenarios and provide convenience
for various research works. Both the size of the datasets and the amount of knowledge contained in the text can reveal the diversity of the benchmark.

\textbf{Representative and useful}
We refine the GeoGLUE benchmark to six typical geographic language processing tasks (e.g., searching, classification, alignment). The content (e.g., datasets, metrics, baselines, etc.) of the GeoGLUE benchmark facilitate further geographic-position-based and geographic-text-based research.

\textbf{Tailor to Geographic-specific characteristics}
Since geographic expressions have the features of the hierarchical location name, ambiguous text, and colloquial phrases, the GeoGLUE benchmark is specially designed to be compatible with the geographic text and text-location relations.

\begin{table*}[tb]
\caption{Train set samples of GeoTES-recall. GeoTES-rerank has a similar train samples format, the negative passages of GeoTES-rerank need to construct by users themselves.}
\centering
\includegraphics[width=0.95\textwidth]{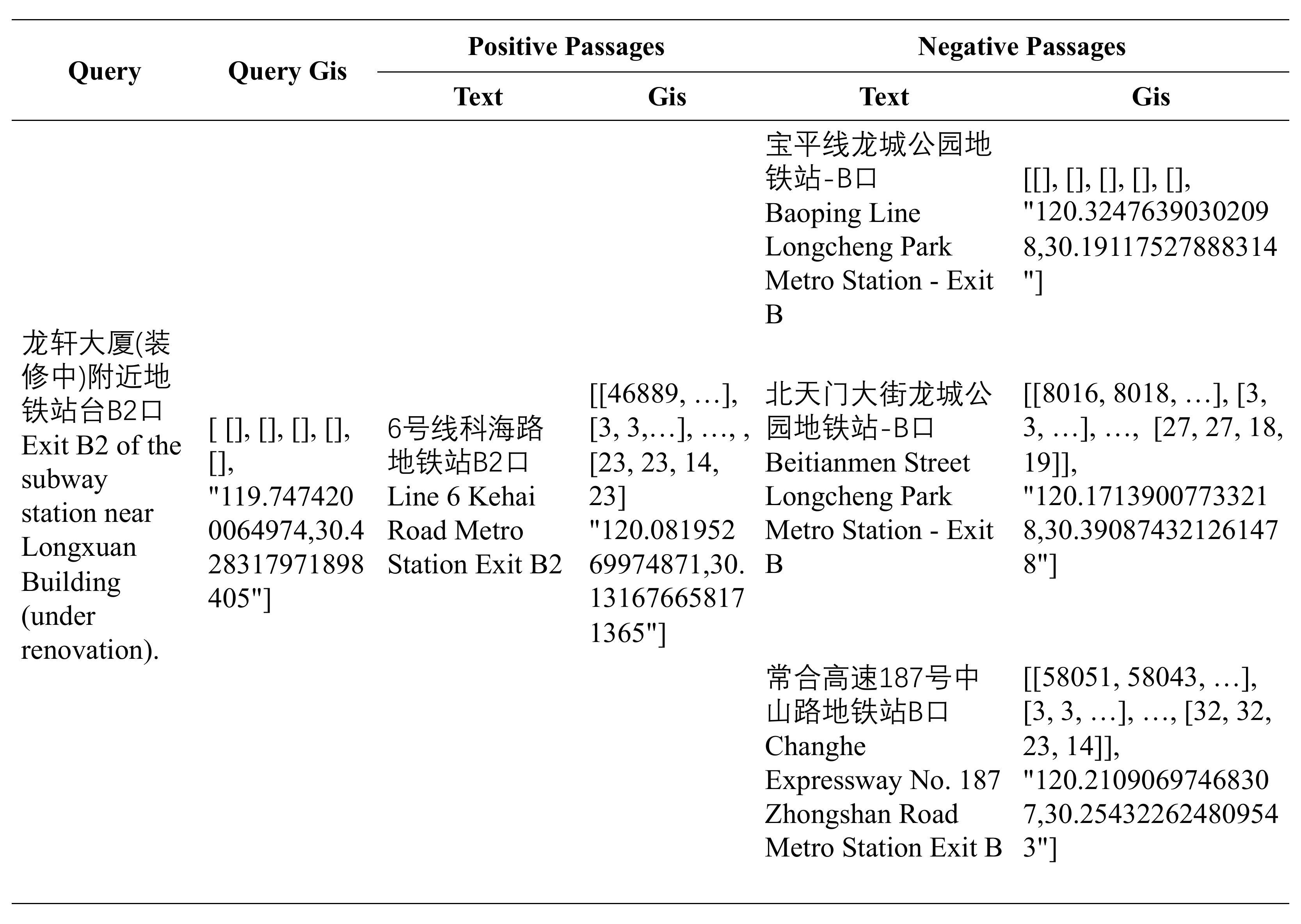}
\label{ranking_trainset}
\end{table*}

\subsection{Tasks}
We classify all GeoGLUE tasks into three categories: Geo-POI Searching, Geo-Sequence Tagging, and Geo-Text Classification.

\subsubsection{Geo-POI Searching}

\begin{itemize}
\item \textbf{GeoTES-recall (Geographic TExtual Similarity on Recall)}
Online map applications (APPs) provide both specific and surrounding location information for searching, positioning, and distance measurement.
POIs retrieval act as an essential role in map APPs' daily business sessions to supply relevant geographic information for the specific query input.
However, the candidate POIs volume is large, and customers' query descriptions contain non-standard and colloquial expressions. 
Geographic TExtual Similarity on Recall task aims to recall the query’s relevant POIs from the POIs database, which preserves volume and various records of POIs. 
Specifically, the recall results should include both POIs descriptions and longitude-latitude descriptions.
We treat MRR@5 as an evaluation metric for this task because we intended to obtain accurate small-range results. Other MRR sorting ranges (i.e., MRR@10, MRR@15, etc.) are also supported.

\item \textbf{GeoTES-rerank (Geographic TExtual Similarity on Rerank)}
When we have recalled a set of relevant POIs from the above GeoTES-recall task, the relevant order of these POIs should also be exactly figured out for further research.
The challenges of this usage scenario are how to distinguish the actual query semantics from the customers' informal expressions and how to provide more accurate candidates order for the results.
Geographic TExtual Similarity on Rerank task aims to rerank the retrieved results by calculating the relevance scores between the query and each candidate.
More relevant candidates should be placed near the front.
Since many geographic applications choose the top-1 result (i.e., GeoCoding) for further operation, we consider MRR@1 as the evaluation metric in this paper. Other MRR sorting ranges (i.e., MRR@5, MRR@10, etc.) are also supported.
\end{itemize}
Note that the recalling task and reranking task share the same training data but different test data, and these two tasks are applied for different retrieval stage models. 
Table \ref{ranking_trainset} lists the train set samples' details, and Table \ref{ranking_results} presents the task results of a specific query example.

\begin{table*}[tb]
\caption{Candidate recall samples of GeoTES-recall. GeoTES-rerank has the same results format.}
\centering
\includegraphics[width=0.95\textwidth]{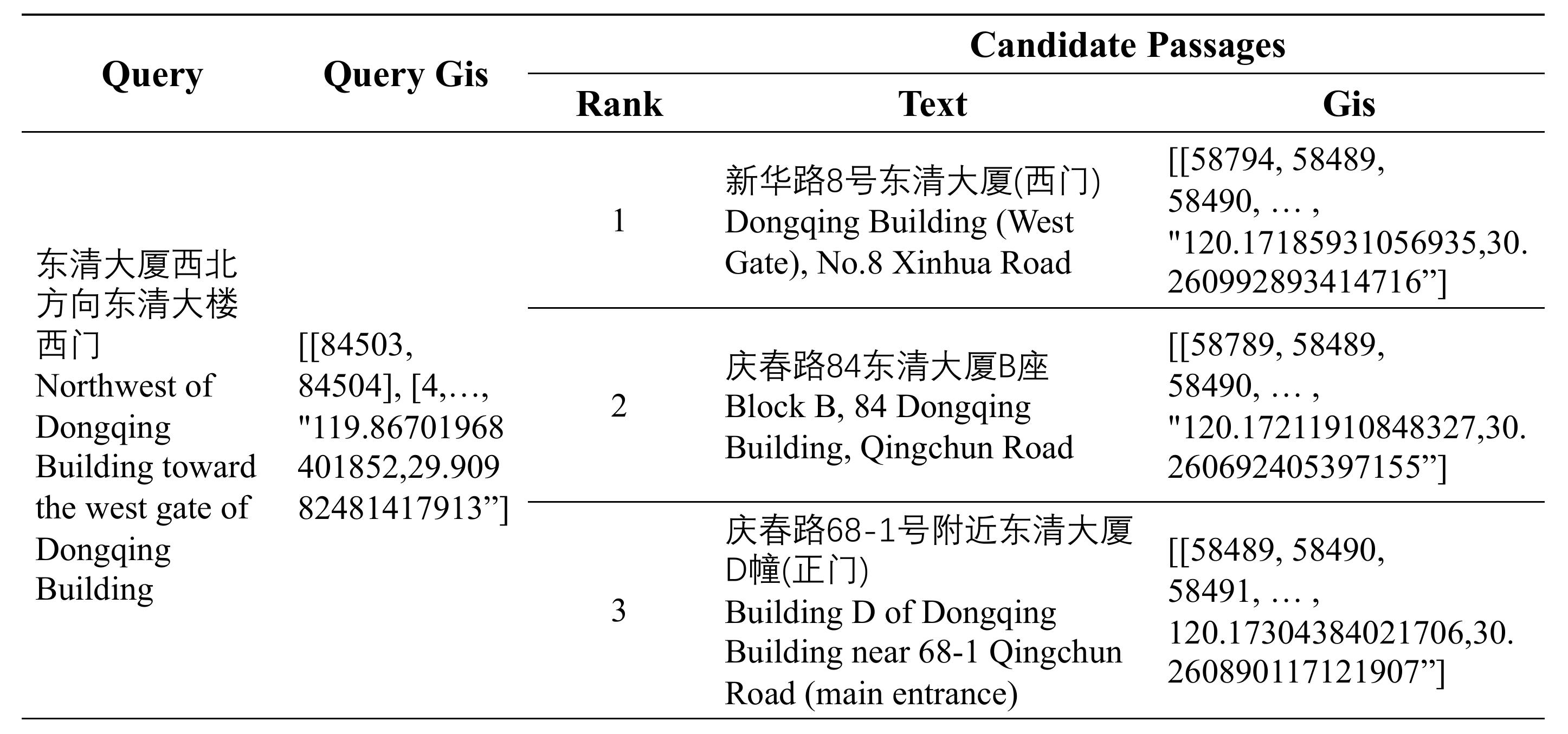}
\label{ranking_results}
\end{table*}

\subsubsection{Geo-Sequence Tagging}

\begin{itemize}
\item \textbf{GeoETA (Geographic Elements TAgging)}
For Logistics Transportation and Take-away delivery fields, room addresses are essential for their business activities. 
Room addresses contain four parts of information: administrative division (e.g., province, city, district, etc.), road network (e.g., road name, road number, etc.), geographic details (e.g., POI, house number, room number, etc.), non-geographic information (e.g., supplements, typos, etc.). 
Recognizing the correct geographic morphemes with the frequently changing location name expression, location abbreviation, and different expressions of the same location are three main challenges of this geographic scenario.
\begin{table*}[!tb]
\caption{Samples of three Geo-Sequence Tagging tasks. Tokens column indicates the first split step, and NER Tags column presents the final tagging results.}
\centering
\includegraphics[width=0.95\textwidth]{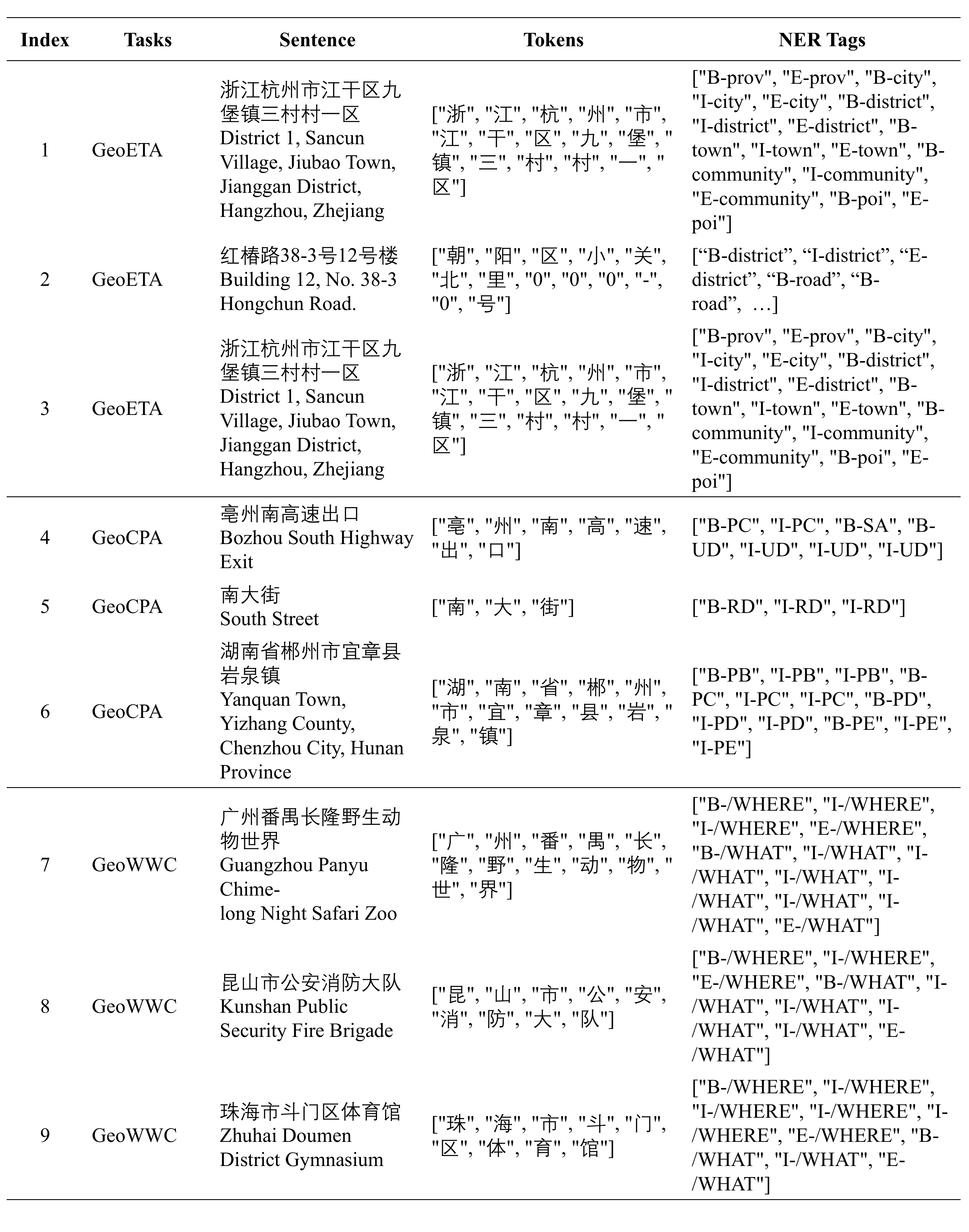}
\label{Tagging}
\end{table*}
The geographic Elements TAgging task attempts to split the input text into independent geographic morphemes and tags each morpheme with the correct label. 
There are 18 categories of morphemes: province, city, district, town, road, road\_number, poi, house\_number, other, etc.
The evaluation metric of this task is precision, recall, and Micro-F1.

\item \textbf{GeoCPA (Geographic ComPosition Analysis)}
In order to satisfy customers' searching purposes (e.g., public transport route information, trading area information, etc.), we need to recognize the detailed components of the query text. 
However, customers' search queries of online map APPs are not always expressed formally. 
Compared with the dataset of GeoETA, this task's input data is less standard and shorter, it also contains brand words and transportation facilities in some cases, which increases the tagging difficulty.
Geographic ComPosition Analysis task is defined to analyze component types of the input text accurately and classifies the components into 28 categories: PC (cities), RD (roads, tunnels, bridges, overpasses), Brand (prominent brands), etc.
The evaluation metric of this task is precision, recall, and Micro-F1.

\item \textbf{GeoWWC (Geographic Where What Cut)}
Address-based queries contain multiple elements, and splitting the integrated queries into separate components is valuable for further utilizations.  
In geographic scenarios, where and what components of the queries are essential and informative, but it is hard to split them exactly, as the inputs contains difficult discriminating location names, mixed expressions of where-what parts, and in-abundant context-aware information.
Geographic Where What Cut task intends to split the where components and what components of address queries separately and tags them with correct labels.
We also exploit precision, recall, and Micro-F1 to evaluate the task results.
\end{itemize}

The splitting and tagging process details are shown in Table \ref{Tagging}.

\subsubsection{Geo-Text Classification}

\begin{itemize}
\item \textbf{GeoEAG (Geographic Entity AliGnment)}
Each geographic language record represents a specific actual location.
Different expressions of the same location exist in different address systems. Reorganizing and aligning different expressions to the correct location are challenging and necessary.
Formal geographic language texts contain four parts of essential information, as introduced in GeoETA task. 
While informal geographic language texts miss some parts and involve colloquial descriptions. 
Geographic Entity AliGnment task seeks to determine whether the two input texts refer to the same location under a predefined rule, especially from some unclear descriptions. 
The alignment results are divided into three types, exact\_match, partial\_match, and not\_match.
We choose Macro-F1 as the task evaluation metric. 
\end{itemize}

Table \ref{alignment} presents three different classifying examples of GeoEAG task. 

\begin{table*}[!tb]
\caption{Three different labeling examples of GeoEAG.}
\centering
\includegraphics[width=0.95\textwidth]{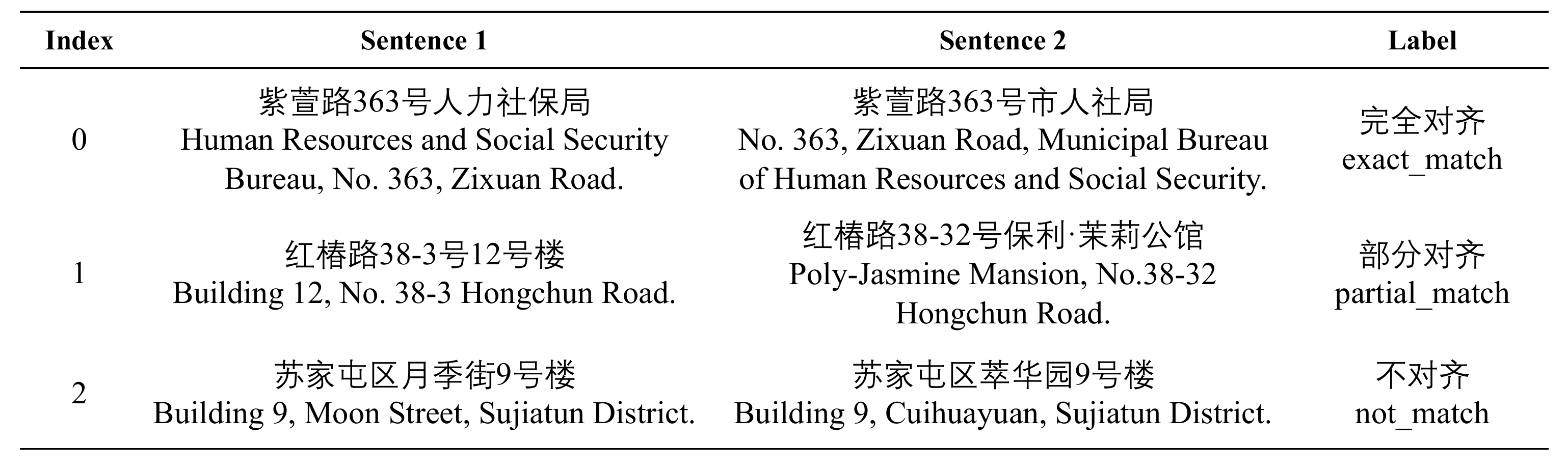}
\label{alignment}
\end{table*}

We list several commonly used metrics for thses six tasks in Table \ref{metrics}, other related appropriate metrics are also supported.
\begin{table}[tb]
\center
\caption{General Evaluation metrics of six tasks.}
\label{metrics}
\begin{tabular}{cc}
\hline
\textbf{Tasks} & \multicolumn{1}{l}{\textbf{ \qquad  General Evaluation Metrics}} \\ \hline
\qquad GeoTES-recall  & MRR@x                                                   \\
\qquad GeoTES-rerank  & MRR@x                                                   \\
\qquad GeoETA         & P, R, Micro-F1                                          \\
\qquad GeoCPA         & P, R, Micro-F1                                          \\
\qquad GeoWWC         & P, R, Micro-F1                                          \\
\qquad GeoEAG         & P, R, Macro-F1                                          \\ \hline
\end{tabular}
\end{table}

\subsection{Data Collection}

\begin{table}[!tb]
\caption{Statistics and metrics of six datasets. ``\#'' means the number of the dataset.}
\center
\begin{tabular}{cc|cccc}
\hline
\multicolumn{2}{c|}{\textbf{Tasks}}                    & \textbf{\#Train} & \textbf{\#Dev} & \textbf{\#Test} & \multicolumn{1}{l}{\textbf{\#Label Types}} \\ \hline
\multirow{2}{*}{Geo-POI Searching}    & GeoTES-recall & 50000             & 20000           & 30002            & -                                           \\
                                      & GeoTES-rerank & 50000             & 20000           & 30002            & -                                           \\
\multirow{3}{*}{Geo-Sequence Tagging} & GeoETA        & 8856              & 1970            & 50000            & 18                                          \\
                                      & GeoCPA        & 49784             & 2214            & 50001            & 28                                          \\
                                      & GeoWWC        & 9311              & 1597            & 50002            & 3                                           \\
Geo-Text Classification               & GeoEAG        & 96423             & 22974           & 35908            & 3                                           \\ \hline
\end{tabular}
\label{Statistics-of-six-datasets}
\end{table}
The detailed statistics of six datasets are shown in Table \ref{Statistics-of-six-datasets}. Only the train and dev set will be released, the test set will be kept on Leaderboard and will not be released to public.
We prepare different datasets for the six natural language processing tasks respectively\footnote{The datasets follow the protocol of \href{https://creativecommons.org/licenses/by-nc/4.0/}{CC BY-NC 4.0}}. All datasets are crawled from online open-released resources, including GIS-based Resources, Secondary-house Trading Resources, Local-information Services Resources, and Query-sessions of Map-related applications. In order to filter the privacy information and obey the anonymity protection rules, we devote additional efforts to refine the original data.

\subsubsection{Collection from GIS-based Resources}
For Geo-POI Searching tasks, we crawl related-POI samples from OpenStreetMap\footnote{\url{https://www.openstreetmap.org}} to obtain GIS-based information for the GeoTES-recall dataset and GeoTES-rerank dataset. 
All GIS-based information is composed of six fields involving geometry ID, geometry type, geometry relationship, geometry corner location, geometry corner ratio of POI longitude-latitude, and POI longitude-latitude. (e.g., [[85462, 85463, 85468, 85469, 85472, 85473], [4, 4, 4, 4, 4, 4], [4, 4, 4, 4, 4, 4], [[622, 736, 626, 764], [622, 736, 626, 764], [615, 762, 626, 766], [615, 762, 626, 766], [626, 764, 631, 770], [626, 764, 631, 770]], [[33, 33, 23, 33], [33, 33, 23, 33], [33, 30, 23, 16], [33, 30, 23, 16], [23, 23, 14, 23], [23, 23, 14, 23]], "119.06179021286454,29.638503988733607"]). In these two Geo-POI searching tasks, queries are manually generated from domain experts who have the geographic background and information intelligence background knowledge. We also execute inspection procedures to provide more accurate samples. 

\subsubsection{Collection from Secondary-house Trading Resources}
For the GeoETA task, we need to collect detailed address-based text and the text should contain administrative division, POI name, and house number. We crawl address information from four commonly viewed websites\footnote{\url{https://www.lianjia.com/}} \footnote{\url{https://www.5i5j.com/}} \footnote{\url{https://www.fang.com/}} \footnote{\url{https://www.mayanchina.com/}}
, only focus on the addresses containing detailed information and filter out the addresses expressions which are non-standard and fuzzy. After refining the data, we split all sentences into tokens to adapt to further processing steps. (e.g., \begin{CJK*}{UTF8}{gbsn}["浙", "江", "杭", "州", "市", "江", "干", "区", "九", "堡", "镇", "三", "村", "村", "一", "区"]\end{CJK*}, "District 1, Sancun Village, Jiubao Town, Jianggan District, Hangzhou, Zhejiang") 

\subsubsection{Collection from Query-sessions of Map-related applications}
For the GeoCPA task and GeoWWC task, we attract data from the query sessions of map-related applications. These two datasets are different from the former GeoETA dataset, GeoETA dataset only focus on detail specific room addresses, while these two datasets are map APPs' query text which contains not only room address but also searching questions and colloquial location names. These records are generated by annotators' imitation of actual scene based on various open-released geographic locations. (e.g., \begin{CJK*}{UTF8}{gbsn}["伊", "宁", "的", "奶", "皮", "子", "什", "么", "地", "方", "有", "卖", "的"]\end{CJK*}, "Where to buy urum of Yining"). We utilize these expressions to support further geographic essential components analysis.

\subsubsection{Collection from Local-information Services Resources}
For the GeoEAG task, we search geographic expressions from local-information services resources. We prepare top-5 related samples to construct five sentence pairs and shuffle them to ensure uniformity. We filter out sentence pairs if two sentences have large length discrepancies. (e.g., \begin{CJK*}{UTF8}{gbsn}"sentence1": "海尔路199号海语江山16栋 Building 16, Haiyu Jiangshan, No.199 Haier Road", "sentence2": "江北区阳明山水16栋 Building 16, Yangmingshanshui, Jiangbei District"\end{CJK*}).

\subsection{Annotation}
All the GeoGLUE tasks' datasets are annotated by experienced annotators who have been trained with corresponding business rules\footnote{Annotators are paid with appropriate compensations, while the specific amounts are commercial confidentialities and will not be disclosed.}. We evaluate the annotated results by different annotators and reach favorable Fleiss' Kappa scores.

For Geo-POI Searching tasks, all POIs come from the WGS84 system\footnote{\url{https://wiki.openstreetmap.org/wiki/Converting\_to\_WGS84/}}. To guarantee the high relevance of related POIs, annotators only select POIs within 1km. The same POIs are annotated as positive passages, and neighboring POIs are annotated as hard negatives, which are helpful to lead the models to promote the distinguishing abilities. Other randomly sampled POIs are treated as normal negatives.

For Geo-Sequence Tagging tasks, we hire 20 annotators to utilize BIOES scheme to tag the samples. They need to split the sentence at the right place and tag different geographic morphemes with correct labels. Each dataset's label statistics is listed in Table \ref{Statistics-of-six-datasets}. When at least 70\% of the grouped annotators generate the same tagging result, the sample is considered correctly tagged.

For Geo-Text Classification task, annotators assign three types of labels to the input sentences. If two sentences indicate the exact same POI, we assign ``exact\_match'' label for the sentence pair. If the two sentences' difference only exists in the last address partition, the sentence pair will be tagged with ``partial\_match''. When two sentences represent two totally different POIs, the pair will be labeled as ``not\_match''. Only at least 70\% of the grouped annotators generate the same tagging result, the sample is considered correctly tagged.

\subsection{Leaderboard}
In order to evaluate the effectiveness of the user's models and produce fair performance scores, we conduct a GeoGLUE Leaderboard\footnote{The Leaderboard site will be released soon.}. In order to ensure the correctness, robustness, and degree of test difficulty, each task's test set consists of ten thousand grades of no-released records.

\section{Experiments}

\subsection{Baselines}
\label{Baselines_section}
We conduct experiments with several general pre-trained language models on GeoGLUE benchmark. Since BERT, RoBERTa, and ERNIE are common and typical pre-trained language models, we provide these baselines for further research scenarios' references. Meanwhile, Nezha is also a typical Chinese pre-train language model and achieves high performance on general Chinese tasks. We choose StructBERT as one of baselines because it is capable of handling long sequence text, and it is an element-order robust model for non-standard colloquial expressions.
\begin{itemize}
\item \textbf{BERT} \cite{DBLP:conf/naacl/DevlinCLT19} is a bio-directional Transformers-based encoder, which utilizes mask language model pre-train objective and next sentence prediction objective to promote the representation produce ability.
\item \textbf{RoBERTa} \cite{DBLP:journals/corr/abs-1907-11692} is an efficient and robust language model developed from BERT, which updates the masking mechanism to a dynamic way and trained on more long sequences. 
\item \textbf{ERNIE} \cite{DBLP:conf/acl/ZhangHLJSL19} proposes an enhanced language model pre-trained on large-scale textual corpora and KGs to fully utilize the lexical, syntactic knowledge.
\item \textbf{Nezha} \cite{DBLP:journals/corr/abs-1909-00204} utilizes whole word masking strategy, functional relative positional encoding, and the LAMB optimizer to train the model.
\item \textbf{StructBERT} \cite{DBLP:conf/iclr/0225BYWXBPS20} proposes a model trained on word level and sentence level objective to take advantage of intra-sentence and inter-sentence relations. 
\end{itemize}

\subsection{Experiments Settings}
All of the experiment results are the average value of 5 times evaluations. We exploit the base-level models in our experiments. The dimension of all models' hidden states is set to 768. All models are set to 12 Transformer layers and 12 self-attention heads. We list hyper-parameters of each model in the order of appearance of baselines in Section \ref{Baselines_section}. The models' learning rates are assigned to \{3e-5, 1e-4, 3e-5, 2e-4, 1e-4\}. We choose \{Adam, Adam, AdamW, AdamW, Adam\} as the optimizer for each model with a learning rate of \{1e-4, 1e-6, 2e-5, 1e-3, 2e-5\} respectively. Weight decay is set to 0.01 for all five models.



\begin{table*}[!tb]
\caption{Experiments performance of different baselines on GeoGLUE benchmark.}
\center
\resizebox{\linewidth}{!}{
\begin{tabular}{ccccccc}
\hline
\multirow{2}{*}{Tasks} & \multicolumn{2}{c}{Geo-POI Searching}       & \multicolumn{3}{c}{Geo-Sequence Tagging}                            & Geo-Text Classification \\ \cline{2-7} 
                        & GeoTES-recall        & GeoTES-rerank        & GeoETA               & GeoCPA               & GeoWWC               & GeoEAG                  \\ \hline
Metrics                 & MRR@5                & MRR@1                & Micro-F1             & Micro-F1             & Micro-F1             & Macro-F1                \\ \hline
BERT                    & 24.59                & 81.52                & 90.41                & 65.06                & 69.65                & 78.88                   \\
RoBERTa                 & 39.37                & 83.20                & 90.79                & 64.65                & 66.44                & 78.84                   \\
ERNIE                   & 24.03                & 81.82                & 90.63                & 66.48                & 67.73                & 79.44                   \\
Nezha                   & 35.31                & 81.38                & 91.17                & {\ul \textbf{67.70}} & {\ul \textbf{70.13}} & {\ul \textbf{79.77}}    \\
StructBERT              & {\ul \textbf{43.10}} & {\ul \textbf{83.51}} & {\ul \textbf{91.22}} & 67.47                & 68.74                & 78.83                   \\ \hline
\end{tabular}}
\label{Experiments-performanc}
\end{table*}

\subsection{Benchmark Results}
We present the performance of five baselines in Table \ref{Experiments-performanc} on the GeoGLUE benchmark. 
From the results, we can observe that 
(1) Although Nezha and StructBERT achieve the best performance on GeoCPA, GeoWWC, and GeoTES-recall task, the results scores are suboptimal in a global perspective. Furthermore, pursuing higher GeoTES-recall scores will assist the GeoTES-rerank task in reaching more accurate results. This suboptimal performance indicates that the GeoGLUE benchmark is challenging and only carefully choreographed models could acquire better performance.  
(2) In five general-domain baseline models, the performance of RoBERTa and StructBERT are significantly higher than others on GeoTES-recall and GeoTES-rerank tasks. RoBERTa's results are  39.37 (MRR@5) and 83.20 (MRR@1), while StructBERT reaches 43.10 (MRR@5) and 83.51(MRR@1). The reason behind this phenomenon is that RoBERTa and StructBERT are trained from long-sequence text and inter-sentence relations, which promotes the multi sentences input and long query-passage style tasks.
(3) Compared with other general models, Nezha obtains the best results on three entity-related tasks, achieving 67.70 on GeoCPA task, 70.13 on GeoWWC task, and 79.77 on GeoEAG task. Nezha's performance demonstrates that its whole word masking mechanism facilitates the ability to learn more effective entity representations, while this capability is relatively weak for the other four models.
(4) ERNIE's results can slightly outperform BERT in most tasks, (90.41 $\mapsto$ 90.63) in GeoETA, (78.88 $\mapsto$ 79.44) in GeoEAG, (81.52 $\mapsto$ 81.82) in GeoTES-rerank, and (65.06 $\mapsto$ 66.48) in GeoCPA. This situation is probably due to the multi-level masking strategy of ERNIE during its training period, while BERT only masks the tokens in a coarse and random way.



\section{Discussions}
We discuss the limitations and future work of our work in this section.
GeoGLUE contains three types of natural language processing tasks; more tasks should be put into practice.
So far, GeoGLUE focuses on Chinese geographic usage scenarios, various languages and multi-lingual geographic text should also be considered.
GeoGLUE is the first work in the geographic natural language understanding evaluation field, and we expect industry and academic colleagues to join us to develop GeoGLUE benchmark together.

\section{Conclusion}

In this paper, we propose a GeoGraphic Language Understanding Evaluation benchmark, named GeoGLUE. GeoGLUE consists of six tasks, geographic textual similarity on recall, geographic textual similarity on rerank, geographic elements tagging, geographic composition analysis, geographic where what cut, and geographic entity alignment. All tasks' datasets are collected from open-released resources and annotated with apposite labels. Experiments on five general baselines demonstrate the significance and challenges of our GeoGLUE benchmark.

\bibliographystyle{splncs04}
\bibliography{reference}
%




\end{document}